# DeepSeek Performs Better Than Other Large Language Models in Periodontal Cases

Hexian Zhang[1], Xinyu Yan[2], Yanqi Yang[2], Lijian Jin[3], Ping Yang[4], Junwen Wang[1,*]

[1]Division of Applied Oral Sciences & Community Dental Care, Faculty of Dentistry, The University of Hong Kong, 34 Hospital Road, Hong Kong SAR, China.
[2]Division of Paediatric Dentistry & Orthodontics, Faculty of Dentistry, The University of Hong Kong, 34 Hospital Road, Hong Kong SAR, China.
[3]Division of Periodontology & Implant Dentistry, Faculty of Dentistry, The University of Hong Kong, 34 Hospital Road, Hong Kong SAR, China.
[4]Division of Epidemiology, Department of Quantitative Health Sciences, Mayo Clinic, Scottsdale, AZ 85259, USA

Hexian Zhang: chordzhang@connect.hku.hk, PhD student, HKU
Xinyu Yan: xinyuyan@connect.hku.hk, PhD student, HKU
Yanqi Yang: yangyanq@hku.hk, Professor, HKU
Lijian Jin: ljjin@hku.hk, Professor and acting dean, HKU
Ping Yang: yang.ping@mayo.edu, Professor, HKU
Junwen Wang: junwen@hku.hk or wang.junwen@mayo.edu, Professor, HKU and Affiliate Professor, Mayo Clinic Scottsdale

*: Corresponds send to Prof Junwen Wang, junwen@hku.hk, Room 2A07, 2/F, PPDH, 34 Hospital Road, Hong Kong SAR, China. Tel: (852) 2852 0128, Fax: (852) 2548 9464

Abstract

Aims: Periodontology, with its wealth of structured clinical data, offers an ideal setting to evaluate the reasoning abilities of large language models (LLMs). This study aims to assess four LLMs (GPT-4o, Gemini 2.0 Flash, Copilot, and DeepSeek V3) in interpreting longitudinal periodontal case vignettes through open-ended tasks.

Materials and Methods: Thirty-four standardized longitudinal periodontal case vignettes were curated, generating 258 open-ended question–answer pairs. Each model was prompted to review case details and produce responses. Performance was evaluated using automated metrics (faithfulness, answer relevancy, readability) and blinded assessments by licensed dentists on a five-point Likert scale.

Results: DeepSeek V3 achieved the highest median faithfulness score (0.528), outperforming GPT-4o (0.457), Gemini 2.0 Flash (0.421), and Copilot (0.367). Expert ratings also favored DeepSeek V3, with a median clinical-accuracy score of 4.5/5 versus 4.0/5 for the other models. In answer relevancy and readability, DeepSeek V3 demonstrated comparable or superior performance, apart from slightly lower readability than Copilot.

Conclusions: DeepSeek V3 exhibited superior reasoning in periodontal case analysis. Its performance and open-source nature support its integration into dental education and research, underscoring its potential as a domain-specific clinical tool.

Introduction

Large language models (LLMs), based on the Transformer architecture, are trained on massive datasets to understand and generate human language(Ouyang et al. 2022). These neural network architectures have proven effective in various natural language processing (NLP) tasks, such as translation and textual analysis. The rapid advancements in LLMs reflect concurrent progress in computing power, data availability the abundance of online data, and algorithmic innovations(Vaswani et al. 2017). Contemporary implementations include OpenAI's GPT series, Google's GEMINI, and Microsoft's Copilot. In 2025, DeepSeek, a large language model developed in China, rapidly gained popularity thanks to its cost-effectiveness, excellent performance, and open-source code. Despite architectural variations in pretraining approaches, they share core capabilities in natural language understanding and generation after iterative training(Silhadi et al. 2025). This technological progress has catalyzed innovation across multiple domains including education and healthcare

In clinical medicine, LLMs are demonstrating potential across multiple applications. For instance, they can aid clinicians through rapid analysis of medical records, patient screening, and clinical documentation support(Jin et al. 2024; Lee et al. 2025). By correlating medical literature with real-world patient data, these models enable doctors to access current evidence and contextualize historical data from clinically similar cases, potentially improving treatment outcomes(Van Veen et al. 2024). Medical education paradigms are similarly evolving through AI-powered intelligent tools like virtual case analysis assistants, highlighting the expanding role of LLMs in healthcare training. Despite these advancements, the integration of LLMs into medical practice remains preliminary(Bedi et al. 2024). Early research primarily assessed LLMs through questions sourced from medical licensing examinations and online databases(Herrmann-Werner et al. 2024), while later investigations incorporated true-false assessments and basic open-ended queries to evaluate foundational reasoning skills(Taymour et al. 2025; Zhou et al. 2023). Initial concerns regarding model-generated inaccuracies, termed "hallucinations," prompted cautious adoption(Farquhar et al. 2024). However, emerging evidence suggests LLMs can now address more complex, open-ended clinical questions with improved reliability, justifying deeper exploration of their capabilities(Goh et al. 2025; Han et al. 2024).

While sharing similar characteristics with clinical medicine, dentistry presents unique operational contexts: most procedures occur in outpatient settings, generating concise, systematically organized case data compared to the heterogeneous records typical of inpatient care(Galler et al. 2016; Laudenbach and Simon 2014). The extended patient-dentist interactions characteristic of dental care foster stronger therapeutic alliances, potentially streamlining processes for data sharing consent and longitudinal case tracking(Damoulis and Gagari 2000). Furthermore, standardized diagnostic criteria and treatment protocols in dentistry reduce ambiguity, offering a structured framework for LLMs to analyze clinical workflows. These features position dentistry as an ideal testing ground for developing AI-assisted decision support systems(Huang et al. 2023).

Current studies on dental LLM applications remain narrowly scoped, frequently focusing on analyzing relatively straightforward data such as multiple-choice questions(Sabri et al. 2025), while overlooking the complex integration of medical histories, diagnostic findings, and phased treatment protocols essential in periodontal management (Hu et al. 2024). This oversight is clinically significant, as approximately half of decision-making errors in periodontal care originate from inadequate synthesis of multisource patient information (Tokede et al. 2024). Our study directly addresses this gap by evaluating LLMs' capacity to interpret longitudinal dental case narratives—a critical skill for replicating clinical reasoning.

Specifically, we investigate whether state-of-the-art models can comprehend detailed case vignettes and generate professionally appropriate responses to open-ended questions. Successful implementation could establish LLMs as simulation platforms for clinical training, equipping dental students and early-career practitioners with AI-augmented tools to study. Beyond education, this work holds implications for enhancing real-world clinical workflows through intelligent case analysis systems.

Methods

The study's methodological workflow was visually outlined in Figure 1.

Data Collection

The data for this study were sourced from the Clinical Case Series published by Wiley-Blackwell(Karimbux 2022), a comprehensive repository detailing real-world clinical workflows. This series encompasses medical histories, diagnostic examinations, treatment planning and implementation, and case discussions. A distinguishing feature of this series is the integration of summarized auxiliary diagnostic results within the case narratives, providing a holistic view of clinical decision-making. To ensure the robustness of our analysis, we selected the Clinical Cases in Periodontics collection as our test corpus. Periodontal case documentation is inherently complex, presenting a greater challenge for large language models (LLMs) in terms of learning and comprehension. From this collection, we extracted 34 cases, each containing between 4 and 12 case-related questions and answers, resulting in a total of 258 question-answer pairs. Ethical review and consent were not required for this study, as all data were derived from previously published research. We secured explicit permission from the publisher, John Wiley and Sons, through License Number 6026290392951 for the reuse of the data.

Data Processing

While modern LLMs support multimodal input, their ability to interpret specialized medical imagery remains suboptimal compared to dedicated vision models, particularly when integrating visual data with extensive textual sequences. Consequently, our study focused exclusively on textual inputs. We designed a three-step conversational framework to structure the interactions with the LLMs. First, the model was assigned a specific role through the prompt: "You are a periodontal tutor. Your role is to provide accurate, clear, and professional answers to periodontal-related questions. When I first input a case or background information, you should only acknowledge it as the context for future questions without summarizing, analyzing, or answering. All subsequent inputs will be specific questions related to the case, to which you should respond directly without revisiting the case unless explicitly asked." Second, the full case narrative was provided to establish context. Third, open-ended clinical questions were presented to the model. All 258 questions were reformatted according to this protocol. To streamline the evaluation process, a random subset of 30% (78 questions) was selected for model testing.

LLM Response

We evaluated four prominent LLMs: OpenAI's GPT-4o, Google's Gemini 2.0 Flash, Microsoft's Copilot, and DeepSeek V3. To mitigate performance variability arising from potential online parameter updates, all model responses for a given question were generated within a one-week window. For consistency, we exclusively utilized non-reasoning-optimized generative models. A new session was initiated for each query, adhering to the prescribed three-step protocol. To assess the possibility of data contamination, we randomly selected 10 cases and input the first half of each vignette into the LLM, instructing it to generate the remaining half portion. The model-generated content was then compared to the original text using the memorization effects Levenshtein detector (MELD) score, which quantifies similarity through character-level matching ratios. This methodology is described in detail in the referenced work(Nori et al. 2023).

Automated Metric-Based Evaluation

We employed the Retrieval Augmented Generation Assessment (RAGAs) framework to evaluate two critical metrics: faithfulness and answer relevancy(Es et al. 2024). The faithfulness score quantifies the factual consistency between generated responses and reference answers on a scale of 0 to 1. It measures the proportion of claims in the LLM outputs that align with ground-truth claims from standard answers, calculated as:

$$\text{Faithfulness Score} = \frac{\text{Number of claims in the response supported by the retrieved context}}{\text{Total number of claims in the response}}$$

The second metric, answer relevancy, assesses the semantic alignment between responses-generated questions and the original questions using cosine similarity (scale: 0–1). This is computed as:

$$\text{Answer Relevancy} = \frac{1}{N} \sum_{i=1}^{N} \text{cosine similarity}\left(E_{g_i}, E_o\right)$$

$E_{g_i}$ denotes the embedding of each generated question; $E_o$ denotes the embedding of the user input

Due to the stochastic nature of RAGAs` claim extraction process, particularly for ambiguous questions, and the inherent variability that persists even with a temperature setting of 0, we conducted three independent evaluation cycles per question and reported the mean scores. Statistical analyses were applied to compare inter-model differences in both faithfulness and relevancy metrics.

Additionally, we assessed answer readability using the Flesch-Kincaid grade level (Steimetz et al. 2024), which estimates the U.S. school grade required for comprehension. It is calculated as follows:

$$\text{Flesch-Kincaid Grade Level} = 0.39 \times \left( \frac{\text{Total Words}}{\text{Total Sentences}} \right) + 11.8 \times \left( \frac{\text{Total Syllables}}{\text{Total Words}} \right) - 15.59$$

Expert-Based Human Evaluation

The outputs generated by different models were anonymized and blinded for both the reviewers and the analyst. To mitigate memory bias, disordered outputs were paired with their corresponding reference answers and presented together after each question. These answers were independently evaluated by two licensed dentists currently pursuing PhD degrees. Credit was awarded for addressing key points from the reference answers, while deductions were applied for medically inaccurate statements; extraneous content was not penalized. Reviewers received standardized training on the evaluation process and scored responses on a 1–5 scale. Inter-rater reliability was calculated using weighted κ statistics. After confirming acceptable agreement, mean scores for each answer were computed for final analysis, and statistical tests were used to determine significant differences across models.

Results

Our evaluation of potential training data contamination in the four LLMs began with an analysis of their MELD scores. The mean MELD scores were consistently low across all models: 0.26 for GPT, 0.26 for DeepSeek, 0.28 for Gemini, and 0.25 for Copilot. Given that a similarity threshold of ≥0.95 is typically required to indicate possible dataset inclusion, these results suggest a negligible likelihood that the models had prior exposure to the evaluated cases. This implies that the generated responses were primarily derived from the models' learned knowledge rather than direct memorization of training data.

Automated assessment metrics for faithfulness, answer relevance, and readability are presented in Table 1. Due to violations of normality assumptions in at least one group, non-parametric statistical analyses were conducted. The Friedman test detected statistically significant differences among the models ($p < 0.05$), leading to post hoc pairwise comparisons using the Wilcoxon signed-rank test. In terms of faithfulness (Figure 2a), DeepSeek (median = 0.528) demonstrated significantly superior performance compared to GPT (median = 0.402), Gemini (median = 0.457), and Copilot (median = 0.367). Regarding answer relevance (Figure 2b), while all models achieved high median scores close to 0.95, Gemini exhibited a notable discrepancy: approximately one-third of its responses scored below 0.5 (Q1 = 0.324), resulting in significantly lower performance relative to GPT (Q1 = 0.917), DeepSeek (Q1 = 0.926), and Copilot (Q1 = 0.919). Readability assessments (Figure 2c) ranked Copilot's outputs as the most accessible (median grade level = 11.9), followed by DeepSeek (12.8), GPT (12.9), and Gemini (13.1).

Expert evaluations of the LLMs are summarized in Table 2. Inter-rater reliability analysis produced a weighted kappa coefficient of 0.784, indicating strong consensus among evaluators. Non-parametric comparisons of median scores revealed DeepSeek (median = 4.5) as the highest-performing model for dental-related queries (Figure 3), significantly outperforming GPT, Gemini, and Copilot (all medians = 4.0). Importantly, all models exhibited clinically acceptable performance, with mean scores exceeding 3.5 (equivalent to 70/100), confirming that LLM-generated responses were non-inferior to author-provided reference answers.

Discussion

The growing adoption of LLMs in clinical settings had prompted critical evaluations of their ability to interpret nuanced case vignettes and generate contextually accurate responses. In this study, we provided a comprehensive comparative analysis of four state-of-the-art LLMs, assessing their performance on dental case analyses through both quantitative computational metrics and expert evaluations. Our findings revealed that DeepSeek consistently outperformed its counterparts, achieved the highest faithfulness scores, and received the most favorable ratings from dental professionals. These results suggested that LLMs, especially DeepSeek, can serve as effective adjuncts to human expertise, offering substantial potential as decision-support tools in both clinical education and practice.

In terms of faithfulness, a key automated evaluation metric that measures how well generated responses align with reference perspectives while minimizing factual inaccuracies and hallucinations(Li et al. 2025), DeepSeek (median = 0.528) significantly outperformed GPT (median = 0.402), Gemini (median = 0.457), and Copilot (median = 0.367). Mirroring human point-by-point grading against benchmarks, this metric assigns scores based on the proportion of reference-answer claims substantiated by the LLM's auto-generated claims, making it well suited for preliminary LLM performance evaluation. Moreover, human assessments on a five-point Likert scale closely matched this ranking, highlighting the substantial gap between DeepSeek's superior performance and Copilot's lower scores, further confirming the metric's reliability.

Answer relevancy further highlighted performance disparities among the models, which quantifies responses' adherence to core questions while minimizing factually correct but tangential content(Chowdhury et al. 2025). In this measure, DeepSeek maintained the highest mean relevancy score, while GPT and Copilot exhibited statistically comparable performance. Notably, Gemini exhibited a different pattern: approximately one-third of its responses fell below the acceptable threshold, despite achieving reasonable faithfulness. This discrepancy suggests that Gemini is more likely to generate factually consistent yet question-irrelevant content, so double-checking is recommended for complex medical inquiries.

Readability emerged as another critical criterion, as higher textual clarity enhances knowledge absorption and user engagement(Rahsepar 2024). Copilot produced significantly superior performance in this metric, while DeepSeek achieved second place despite typically generating the most extensive responses. Its ability to convey comprehensive content with clarity is particularly noteworthy, as extensive responses often risk sacrificing readability. In contrast, Gemini lagged significantly, showing statistically significant gaps compared to both Copilot and DeepSeek. This hierarchy suggests that output length does not need to compromise readability when optimally structured.

Expert evaluations served as the gold standard for assessing LLM output quality(Huang et al. 2024). After training the evaluators, initial scoring revealed some discrepancies ($\geq 2$ point differences) in 8 questions. To resolve inconsistencies, we had the two dentists independently re-evaluate all four responses from the LLMs to these questions under blind conditions. The resultant weighted kappa coefficient of 0.784 confirmed strong inter-rater reliability. DeepSeek demonstrated overwhelming superiority over other models, whereas Copilot consistently underperformed, a pattern mirroring their faithfulness metric rankings. This alignment suggests faithfulness scores effectively approximate human judgment for primary quality assessment, offering a reliable, cost-effective surrogate for human evaluation. While answer relevancy and readability metrics provide supplementary insights, they should primarily serve as refinements to faithfulness-driven conclusions.

Based on the above evaluation, DeepSeek exhibited exceptional performance on case-related open-ended clinical tasks. This advantage can be attributed to DeepSeek's mixture-of-experts (MoE) architecture, which employs dynamic query routing to specialized neural sub-networks(Liu et al. 2024). Such design empowers the model to more effectively harness domain-specific knowledge, producing responses characterized by both precision and clinical relevance. Notably, as the current state-of-the-art open-source model, DeepSeek's demonstrated performance suggests promising potential for capability enhancement through architectural optimization. Recent clinical study has further substantiated DeepSeek's superior performance(Sandmann et al. 2025), corroborating our conclusions regarding its efficacy in addressing complex challenges within the dental domain.

The study also had several limitations. First, the gold-standard reference answers were unilaterally defined by the authors, which may have introduced bias, particularly for ambiguous questions. This rigidity occasionally led to discrepancies where LLMs generated responses rated by experts as higher quality than the reference answers but were erroneously penalized in automatic metrics. Such cases underscored the need for future work to establish consensus-based gold standards through multi-expert validation. Additionally, image data were excluded due to technical constraints, including challenges in uploading large image files and limitations in storing them within the JSON format. As LLMs continue to advance, however, these barriers may be overcome, enabling more robust analysis of clinical cases, including those lacking figure information, by integrating image inputs directly into initial assessments.

To investigate performance optimization strategies, we conducted fine-tuning experiments on GPT-4o, using 70% of the original data (258 Q&As in total) for training, 20% for validation, and 10% for testing. After hyperparameter optimization, the results showed that the average faithfulness score improved from 0.421 to 0.457. However, expert evaluation scores decreased slightly from 4.19 to 3.96. The fine-tuned model generated shorter answers than the original GPT-4o, with no evidence in all test answers suggesting superior expert scores for the fine-tuned version. This indicates that fine-tuning chiefly optimized answer presentation to strike an ideal balance between conciseness and accuracy, while also fostering a more professional tone, reducing hallucinations, and enhancing overall readability(Wu et al. 2024). The curated dataset is expected to expand medical knowledge coverage for better performance on related tasks in future iterations. Moving forward, we aim to develop a larger domain-specific dataset and build a specialized medical language model using open-source foundations like DeepSeek, fostering broader clinical and educational applications in medicine(Sirrianni et al. 2022).

Conclusion

In this study, we demonstrated that large language models (LLMs) are well suited for answering dental case–vignette questions. Of the four widely recognized LLMs we evaluated, DeepSeek outperformed the others, delivering superior results and establishing itself as the optimal choice for this task. Our findings support integrating LLMs into clinical practice to boost efficiency and accessibility without compromising accuracy. Going forward, we recommend prioritizing the creation of domain-specific LLMs trained on extensive literature and case-based datasets. Such tailored models could further enhance precision, conciseness, and clinical relevance, thereby accelerating the adoption of AI-driven solutions in medicine.

Author contributions

Hexian Zhang contributed to conception and design and drafting of the manuscript; Xinyu Yan contributed to acquisition and interpretation of the data and drafting of the manuscript; Yanqi Yang contributed to analysis of the data and critical revision of the manuscript; Lijian Jin contributed to interpretation of the data and critical revision of the manuscript; Ping Yang contributed to analysis of the data and critical revision of the manuscript; Junwen Wang contributed to conception and critical revision of the manuscript. All authors reviewed and approved the final version of the manuscript and agreed to be accountable for all aspects of the work.

Declaration of Conflicting Interests

The authors declare that there is no conflict of interest regarding the publication of this article.

Funding

The study was funded by collaborative research fund of Hong Kong RGC (C7015-23G), seed funding for collaborative research (2207101590) and basic research (2201101499) from the University of Hong Kong, and startup funds [207051059, 6010309] from Faculty of Dentistry, the University of Hong Kong to J Wang.

Data availability

All data generated or analysed during this study are included in [Extended Data]. You can contact the corresponding author to obtain the file. We developed a numerical case index based on the textbook "Clinical Cases in Periodontics" by Nadeem Karimbux and published by Wiley-Blackwell to organize case vignettes, which were directly input into LLMs for evaluation. The dataset includes specific questions, responses generated by the four LLMs, and evaluation metric scores.

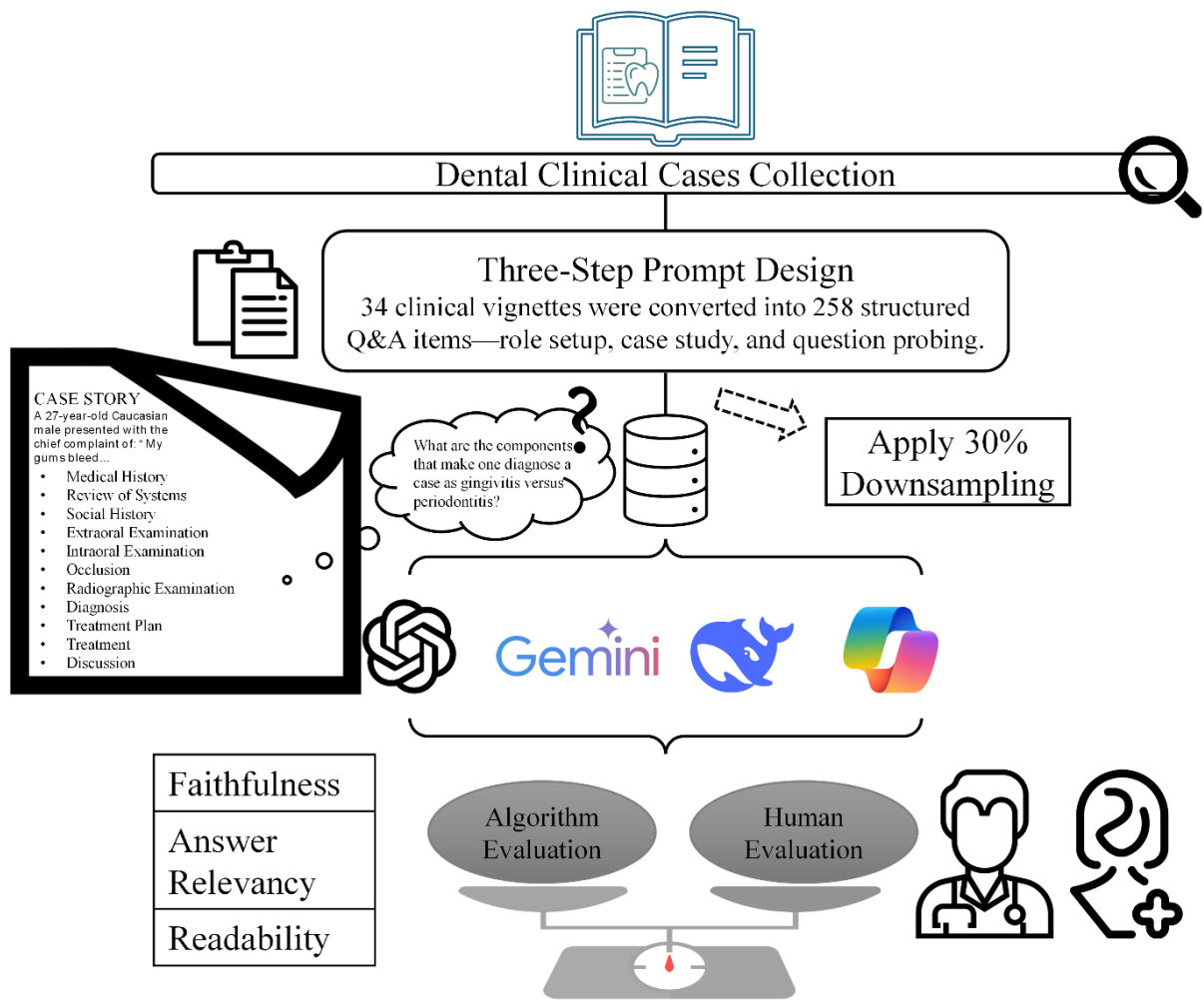

Figure 1: A framework for evaluating the dental question-answering proficiency of LLMs. Thirty-four comprehensive clinical cases (each 2,000–6,000 words, including chief complaints, medical histories, clinical and radiographic examinations, diagnoses, treatment plans, procedures, follow-ups, and discussion sections) were structured into a three-step input format, yielding 258 questions. We randomly selected 30% of these (n = 78) and presented the formatted cases to four major LLMs. Model performance was assessed using automated metrics and expert evaluations.

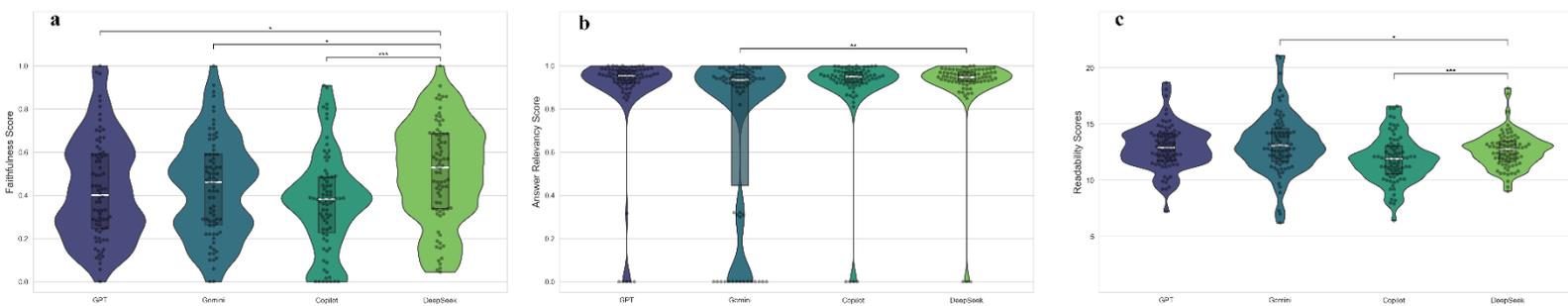

Figure 2: Model performance on automated assessment metrics. (a–c) Violin plots with modified box plots compared the performance of four LLMs on (a) faithfulness scores (GPT vs. DeepSeek,

Mann–Whitney test, p = 0.016; Gemini vs. DeepSeek, p = 0.028; Copilot vs. DeepSeek, p < 0.001), (b) answer relevance scores (Gemini vs. DeepSeek, p = 0.003), and (c) readability grades (Copilot vs. DeepSeek, p < 0.001; Gemini vs. DeepSeek, p = 0.019). Significant differences between DeepSeek and the other models were highlighted.

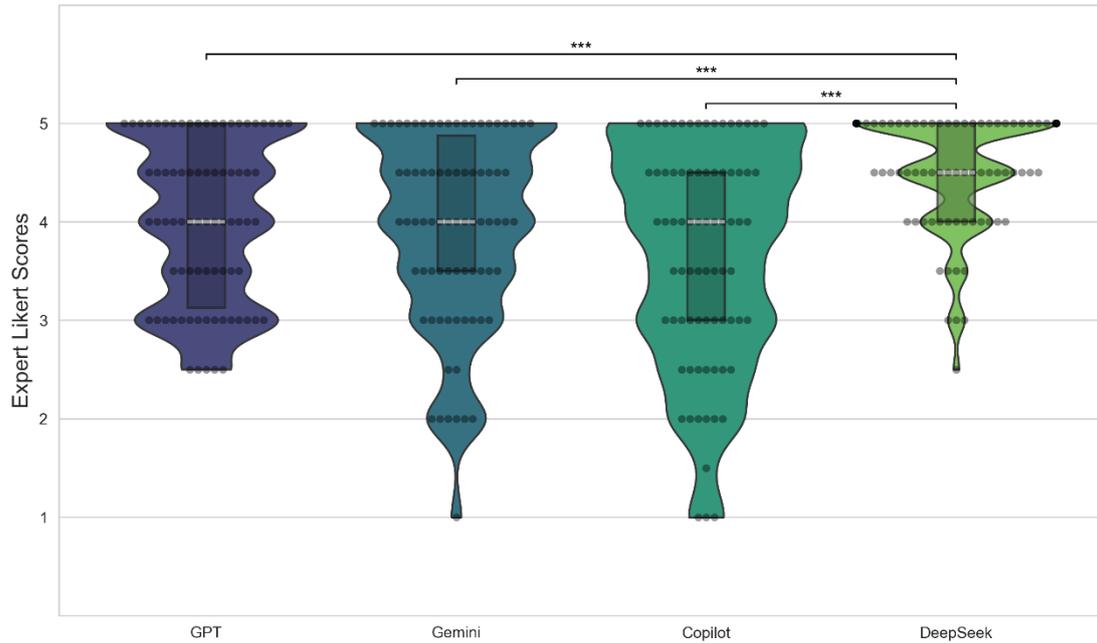

Figure 3: Expert evaluation results on a 5-point Likert scale. Violin plots with modified box plots illustrated the distribution of mean scores assigned by experts to four LLMs. Significant differences between DeepSeek and the other models were observed (Mann–Whitney test, p < 0.001).

| Metric | Model | Mean | Median (IQR) | P-Value |
|---|---|---|---|---|
| Faithfulness | DeepSeek | 0.506 | 0.528 (0.335-0.685) | |
| | GPT | 0.429 | 0.402 (0.235-0.587) | 0.016 <DeepSeek |
| | Gemini | 0.437 | 0.457 (0.256-0.590) | 0.028 <DeepSeek |
| | Copilot | 0.367 | 0.381 (0.220-0.484) | 0.023 < Gemini<br><0.001 <DeepSeek |
| | | | | 0.018 (Overall) |
| Answer Relevancy | DeepSeek | 0.910 | 0.946 (0.926-0.970) | |
| | GPT | 0.880 | 0.952 (0.917-0.979) | |
| | Gemini | 0.723 | 0.935 (0.324-0.962) | 0.003 <DeepSeek<br>0.001 <GPT<br>0.015 <Copilot |
| | Copilot | 0.896 | 0.948 (0.919-0.969) | |
| | | | | 0.039(Overall) |
| Readability | DeepSeek | 12.7 | 12.8 (11.6-13.5) | 0.001 <Copilot |
| | GPT | 12.9 | 12.9 (11.7-14.2) | <0.001 <Copilot |
| | Gemini | 13.3 | 13.1 (11.6-14.7) | 0.019 <DeepSeek<br><0.001 <Copilot |
| | Copilot | 11.8 | 11.9 (10.5-13.1) | |
| | | | | <0.001(Overall) |

Table 1: A comparative analysis of automated evaluation metrics across four LLMs. The P-value elucidates the holistic disparity, with those pairwise post-hoc comparisons achieving statistical significance ($p<0.05$) being displayed ('<' indicating median difference direction). IQR: interquartile range.

|  | Model | Mean | Median (IQR) | P-Value |
|---|---|---|---|---|
| Expert Evaluation | DeepSeek | 4.513 | 4.5 (3.5-5) | |
| | GPT | 4.013 | 4 (3.125-5) | <0.001 <DeepSeek |
| | Gemini | 3.930 | 4 (3.5-4.875) | <0.001 <DeepSeek |
| | Copilot | 3.667 | 4 (3-4.5) | <0.024 <Gemini<br><0.002 <GPT<br><0.001 <DeepSeek<br><0.001(Overall) |

Table 2: Performance comparison of expert evaluations on a 5-point Likert scale across four LLMs. The P-value elucidates the holistic disparity, with those pairwise post-hoc comparisons achieving statistical significance (p<0.05) being displayed ('<' indicating median difference direction). IQR: interquartile range.